\documentclass[10pt,twocolumn,letterpaper]{article}

\usepackage{cvpr}              

\usepackage[misc]{ifsym} 
\usepackage{booktabs}
\usepackage{makecell}
\usepackage{multirow}
\usepackage{xspace}

\definecolor{cvprblue}{rgb}{0.21,0.49,0.74}
\usepackage[pagebackref,breaklinks,colorlinks,allcolors=cvprblue]{hyperref}

\def\paperID{*****} 
\def\confName{CVPR}
\def\confYear{2026}

\newcommand{\tabref}[1]{Tab.~\ref{#1}}
\newcommand{\figref}[1]{Fig.~\ref{#1}}

\newcommand{\methodname}{MeV\xspace}

\title{Mise-en-Video: Fine-Grained 3D Control for Multi-Shot Video Generation}

\author{
Yingshuang Zou\textsuperscript{1$\dagger$}\quad
Xiaoyu Shi\textsuperscript{2}$^\text{\Letter}$\quad
Yawen Luo\textsuperscript{3}\quad
Ziang Li\textsuperscript{4}\quad
Qinghe Wang\textsuperscript{5}\\
Mengfei Li\textsuperscript{1}\quad
Chengfeng Zhao\textsuperscript{1}\quad
Xintao Wang\textsuperscript{2}\quad
Pengfei Wan\textsuperscript{2}\quad
Kun Gai\textsuperscript{2}\quad
Yuan Liu\textsuperscript{1}$^\text{\Letter}$\\[0.5em]
\textsuperscript{1}The Hong Kong University of Science and Technology\quad
\textsuperscript{2}KlingAI Research\\
\textsuperscript{3}The Chinese University of Hong Kong\quad
\textsuperscript{4}Northwestern Polytechnical University\\
\textsuperscript{5}Dalian University of Technology\\[0.25em]
{\tt\small yzoubb@connect.ust.hk, yuanly@ust.hk}
}

\begin{document}

\twocolumn[{
\renewcommand\twocolumn[1][]{#1}
\maketitle
\vspace{-0.9cm}
\begin{center}
    \centering
    \captionsetup{type=figure}
    \includegraphics[width=1\linewidth]{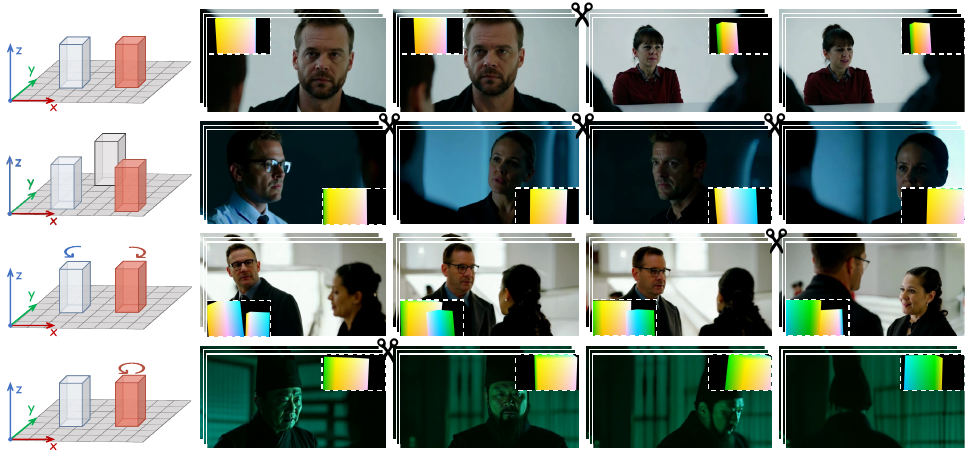}
    \captionof{figure}{We propose \methodname to help creators generate high-quality multi-shot video content. For each shot, creators specify a 3D layout, akin to staging a real 3D set (first column), and define actors' appearance, trajectories, positions, and other 3D scene parameters. We render the 3D layout as a control signal (shown in the insets) and feed it into the generator, enabling stronger cross-shot consistency and finer-grained controllability.}
    \label{fig:teaser}
\end{center}
}]

{
\let\thefootnote\relax\footnotetext{$\dagger$ This work was conducted during the author's internship at KlingAI.}
\let\thefootnote\relax\footnotetext{$\text{\Letter}$ ~Corresponding authors.}
}
\begin{abstract}
Recent advances in text-to-video generation have enabled high-fidelity single-shot synthesis, yet practical storytelling creation requires multi-shot videos with both strong inter-shot consistency and fine-grained controllability.
Existing multi-shot video generation approaches achieve strong consistency but still fall short in controllability.
In this paper, we propose Mise-en-Video (\textbf{\methodname}), a highly controllable multi-shot video generation framework that allows video creators to conduct staging the multi-shot in the 3D space and then generate the corresponding multi-shot video.
To achieve this, our framework combines a hierarchical attention mechanism for multi-level narrative consistency with 3D-aware layout guidance for precise, per-shot spatial control.
To support training and evaluation, we develop standardized annotation pipelines and construct a 160K-video dataset with hierarchical multi-shot texts and oriented 3D bounding boxes.
Extensive experiments demonstrate that \methodname consistently outperforms strong baselines in video quality, inter-shot consistency, and subject-level controllability. Project page: \href{https://heiheishuang.xyz/Mise-en-Video}{\textcolor{magenta}{https://heiheishuang.xyz/Mise-en-Video}}.
\end{abstract}




\section{Introduction}
With the continuous iteration and maturity of text-to-video generation technology~\cite{wan2025wan,kong2024hunyuanvideo,team2025kling}, it has been widely applied in artistic creation, film production, and other fields, and has become an important technical support in the field of content production. 
However, the current mainstream video generation models still have obvious limitations. Although they perform well in synthesizing high-fidelity single-shot videos, they find it difficult to meet the more complex content needs in practical applications. 
In narrative scenarios such as film and television, the presentation of long-form stories often relies on the organic combination of multiple shots~\cite{guo2025long}, which is still a challenging and open problem for current video generation models. 
Especially the difficulty in simultaneously ensuring inter-shot consistency and fine-grained controllability, which has become a key bottleneck restricting the practical application of T2V technology.

Multi-shot video generation puts higher requirements on generation models in two aspects: consistency and controllability. First, to achieve coherent narrative expression, the generated multi-shot videos must maintain consistency in the background environment, core theme, and characters across shots. 
Second, beyond mere narrative stability, professional content creation requires granular steerability over the generative process. As creators seek to move past randomized outputs, the focus has shifted toward precisely manipulating key cinematic elements, such as shot composition, subject positions, and camera angles, making controllable multi-shot generation a pivotal frontier in the field.

Existing multi-shot video generation methods generally follow two paradigms: decoupled generation and one-time generation. Earlier decoupled approaches~\cite{xiao2025captain, zheng2024videogenofthought} generate shots independently and splice them together; while this allows for individual shot control, it often fails to maintain inter-shot consistency in identity and style. Conversely, one-time generation methods~\cite{meng2025holocine, wu2025cinetrans, jia2025moga, cai2025mixtureofcontext} leverage cross-shot attention to ensure global coherence, yet they often trade off fine-grained controllability. Most such models solely rely on text prompts for each shot to guide shot generation, which we observe to be insufficient for complex creation. 


Specifically, our key observation is that such a text attention mechanism ignores the fact that the videos are projections of 3D scenes and video creators (directors) usually first conduct 3D staging to determine the subject layout and then fill in the details for each subject.
In contrast, current multi-shot generation methods directly input all the textual details of each shot and leave the video generation model to determine the subject layout.
This causes the following two issues.
1. Confusion of spatial relationships: Text descriptions are often underspecified or ambiguous regarding 3D geometry, and without explicit 3D modeling, the generated shots may exhibit inconsistent depth ordering and unreasonable spatial layouts.
2. Inability to control dynamic movements: Text prompts naturally lack the spatial awareness, making them hard to specify subject motions such as movement trajectories and turning directions. However, to ensure the correct narrative, video creators have a strong need to specify the movement and actions of actors.

In this paper,  we propose \textit{Mise-en-Video} (\textbf{\methodname}), a novel and highly controllable multi-shot video generation framework, which effectively assists artists in creating high-quality multi-shot videos. Within \methodname, creators can specify the 3D layout of each shot, like setting up a real 3D stage, and define the movement of actors, the position of objects, and other 3D-related parameters. Directly staging the plot in the 3D space not only ensures the correctness of 3D spatial relationships but also enables precisely specifying the dynamics of subjects in the 3D space.

Designing such a 3D-aware multi-shot video generation method is non-trivial. 
Such a framework requires taking both the text prompts and 3D layout as conditions into a 2D video generation, which suffers from severe modality mismatch.
First, the 3D layout designed by the creators needs to be converted into effective 2D control signals for video generation. 
However, fine-grained 3D geometry information, such as point clouds or depth maps, is not only hard to obtain but also lacks flexibility in practical creation. 
Meanwhile, simply adopting coarse 3D bounding boxes cannot accurately control the detailed behaviors of subjects, such as character turning and posture adjustment.
Therefore, it is still a challenge to design a flexible but steerable 2D control from the 3D layout. 
Second, how to effectively associate the subject textual descriptions with the specific 3D subject still remains unexplored. A multi-shot video generation usually involves multiple subjects. Simply associating each video shot with the shot description without subject-level association often leads to confusion about the subject.


To address the above challenges, we adopt two key designs.
First, we design a \textbf{Subject Layout Guidance} to convert 3D layouts into 2D control signals.
Considering the trade-off between control flexibility and representational complexity, we represent each subject using oriented 3D bounding boxes. This representation is intentionally lightweight, allowing creators to intuitively define the 3D layout and blocking of a scene without the need for exhaustive geometric modeling.
To bridge the gap between coarse boxes and fine-grained subject dynamics, we propose a NOCS~\cite{wang2019normalized}-based (Normalized Object Coordinate Space) color encoding scheme. By mapping canonical coordinates onto the 3D boxes, we transform abstract orientations and movements, such as a character turning or tilting, into dense, orientation-aware visual cues.
These 3D representations are projected into the 2D camera view to form per-frame layout maps. To resolve the modality mismatch between 3D-structured signals and 2D pixel generation, we employ a ControlNet-like structure to inject these signals into the Diffusion Transformer (DiT) backbone. 



Then, to effectively associate the subject texts with the subject bounding box, we adopt the \textbf{Hierarchical Context Attention} mechanism. 
To ensure the fluency of multi-shot narrative and the rationality of subject interaction, we divide the text description into three levels: global level, shot level, and subject level. 
Then, we modify the attention weights to restrict different levels of descriptions to different levels of video content. Global texts will be visible to all tokens, shot-level texts will be visible only to the specific shot, and the subject-level texts are restricted to the subject token regions.
The Hierarchical Context Attention mechanism can efficiently capture and fuse information from these three levels, improving the model’s adherence to global narrative intent, shot-specific requirements, and subject-level attributes, thereby enhancing both controllability and cross-shot consistency.

By coupling Hierarchical Context Attention with Subject Layout Guidance, our model effectively utilizes 3D signals to control the multi-shot video generation. To train our model, we construct a 160K video dataset, which includes multi-shot hierarchical text descriptions, NOCS color maps, and 3D bounding boxes with orientation information. This dataset provides solid data support for model training and evaluation, improving the robustness and generalization of the framework.

Extensive experiments validate the effectiveness of the proposed framework. Results show that \methodname consistently outperforms baselines in video quality, inter-shot consistency, and subject-level controllability.
In particular, \methodname achieves improved performance on both general video quality metrics and specialized controllability metrics, demonstrating the ability to support fine-grain control for multi-shot video creation.


\begin{figure*}[t]
  \centering
  \includegraphics[width=0.95\linewidth]{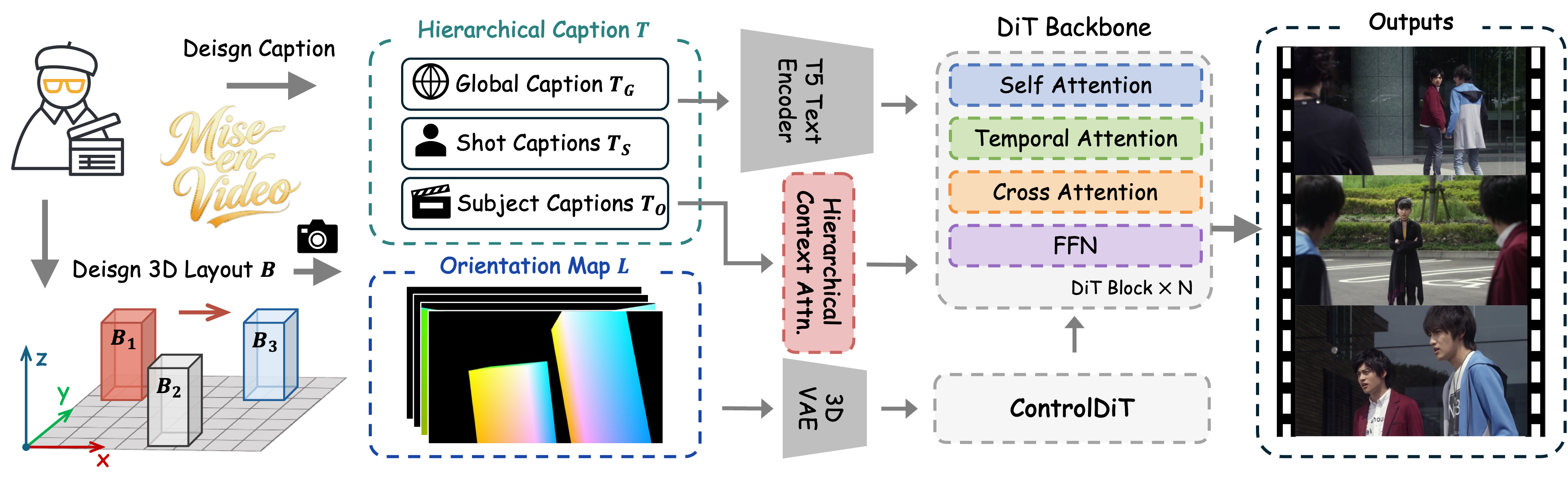}
  \caption{Overview of \methodname. Creators first design the 3D scene and provide a hierarchical caption. We project the 3D layout into an NOCS-based encoding scheme and use it as guidance for ControlDiT. Meanwhile, we decompose the caption into global-, shot-, and subject-level descriptions, and fuse them via a Hierarchical Context Attention mechanism to enhance narrative consistency and subject coherence across shots.}
  \label{fig:pipeline}
\end{figure*}

\section{Related Work}
\subsection{Single-Shot Video Generation}
The advent of Diffusion Transformer~(DiT)~\cite{peebles2023scalable} architectures has brought substantial improvements to the video generation in visual quality, duration, and text prompt following~\cite{videoworldsimulators2024,gao2025seedance,opensora}. The widely-used HunyuanVideo~\cite{kong2024hunyuanvideo} and Wan~\cite{wan2025wan} are constructed by significantly scaling up model capacity and training data, paving the way for the community to delve into fresh frontiers of world modeling and its diverse applications. To further push the boundary of single-shot video length, several works have explored autoregressive and streaming generation paradigms. FramePack~\cite{zhang2025packing} proposes a structured frame packing strategy that compresses historical context into compact representations, allowing the model to generate videos of arbitrary length in a sliding-window fashion while maintaining temporal consistency. Self-Forcing~\cite{huang2026self} addresses the training-inference discrepancy in autoregressive video generation by training the model to condition on its own previously generated frames, enabling stable long-video synthesis without exposure bias. Despite these advances in generating longer and more temporally coherent single-shot videos~\cite{huang2025memory,yang2025longlive}, they remain fundamentally constrained to a single continuous camera shot. Such single-shot video generation lacks narrative expressiveness and is unable to convey rich information through varied shots as films do, making it difficult to captivate audiences for extended viewing.


\subsection{Multi-Shot Video Generation}
With the rapid advancement of single-shot video generation methods~\cite{wan2025wan,team2025kling}, these models have been widely adopted in commercial scenarios such as advertisement production and media content creation. Consequently, the research focus is shifting from single-shot to multi-shot video generation~\cite{guo2025long,jia2025moga,cai2025mixtureofcontext,meng2025holocine,wu2025cinetrans,wang2025echoshot,luo2026shotstream,wang2025multishotmaster}. This emerging paradigm aims to generate videos composed of multiple shots while maintaining subject and scene consistency across varying camera perspectives, thereby facilitating complex narrative storytelling. Existing multi-shot video generation methods generally fall into two categories. The first category~\cite{zheng2024videogenofthought,xiao2025captain} employs a two-stage approach: it initially generates the starting frame of each shot and subsequently uses an image-to-video method to extend them. However, this approach yields suboptimal results, as consistency is only guaranteed across the initial frames. The second category~\cite{meng2025holocine,wu2025cinetrans,wang2025multishotmaster,jia2025moga,cai2025mixtureofcontext} leverages contextual learning to model the relationships between shots, ensuring better global consistency. Although effective, these models~\cite{meng2025holocine,jia2025moga,cai2025mixtureofcontext,wu2025cinetrans} suffer from limited controllability, as they can only guide the generated content through text prompts, making it difficult for users to achieve their precise desired outcomes.

\subsection{Controllable Video Generation}
Precise controllability is essential for user-directed video generation. 
Recent methods control camera motion~\cite{he2024cameractrl,gu2025das,gu2026go,bai2025recammaster,liu2026omnidirector,chen2026anyrecon}, visual style~\cite{ye2025stylemaster,liu2023stylecrafter}, subject identity~\cite{jiang2025vace,huang2025conceptmaster}, object motion in driving scenes~\cite{li2025uniscene,zou2025mudg,zheng2023layoutdiffusion,gao2024magicdrive,guo2025dist,gao2025magicdrive,li2024drivingdiffusion}, and human motion~\cite{hu2024animate,zhu2024champ,zhao2026comovi,cao2025uni3c}.
CameraCtrl~\cite{he2024cameractrl}, CineMaster~\cite{wang2025cinemaster}, and CamCloneMaster~\cite{luo2025camclonemaster} control camera motion, while VACE~\cite{jiang2025vace} and ConceptMaster~\cite{huang2025conceptmaster} preserve subject identity. 
Despite their effectiveness, a major limitation of these approaches is that they are predominantly restricted to single-shot settings.
Driving-oriented methods mainly use layouts or bounding boxes to control object positions and trajectories in structured road scenes, whereas our setting targets 3D subject placement and orientation across cinematic shots. 
To address multi-shot controllability, MultiShotMaster~\cite{wang2025multishotmaster} recently proposed the first controllable multi-shot video generation framework. 
However, its controllability remains confined to the 2D domain, limiting fine-grained control over subject orientation and spatial positioning critical for cinematic storytelling. 
Thus, we propose a fine-grained, 3D-controlled multi-shot video generation model, empowering creators with precise spatial control for complex video production.

\section{Method}

\subsection{Overview}
We propose \methodname for coherent multi-shot storytelling video generation conditioned on hierarchical text and explicit subject-level layout control. 
We first introduce Subject Layout Guidance by encoding each subject with oriented 3D bounding boxes and a NOCS-based representation, projecting them into per-frame 2D layout maps and injecting them into the ControlDiT backbone to precisely control subject placement and facing direction over time (Sec. \ref{subsubsec:guidance}).
We then decompose the caption into global, shot, and subject levels and fuse them with a Hierarchical Context Attention mechanism to improve narrative consistency and subject coherence across shots (Sec. \ref{subsubsec:hca}).
To support large-scale training, we also build a multi-shot video annotation pipeline that provides hierarchical captions and shot-wise subject layout supervision (Sec. \ref{subsec:annotation}).

\subsection{Storytelling Multi-Shot Video Generation}
\subsubsection{Preliminary}
\paragraph{Model.}
Our model is built upon a pretrained single-shot text-to-video (T2V) model with approximately 1B parameters. The backbone consists of a 3D variational autoencoder (VAE), a T5 text encoder, and a latent Diffusion Transformer (DiT). Each basic DiT block comprises a sequence of modules, including a 2D spatial self-attention module, a 3D spatiotemporal self-attention module, a text cross-attention module, and a feed-forward network (FFN). 

\paragraph{Training.}
Given a text prompt, we first feed it into a T5 encoder to obtain the text conditioning feature $c_{\text{text}}$.
For each training video, we encode it into a clean latent $z_0$ using the pretrained 3D VAE.
Following Rectified Flow, we sample $\tau \sim \mathcal{U}(0,1)$ and $z_1 \sim \mathcal{N}(0,\mathbf{I})$, and construct an intermediate latent along the straight path:
\begin{equation}
z_\tau = (1-\tau)z_0 + \tau z_1 .
\end{equation}
We then input $(z_\tau,\tau,c_{\text{text}})$ into the DiT denoising network $v_\Theta(\cdot)$ to predict the velocity field.
The model is trained by regressing the target velocity $(z_1 - z_0)$ with the following objective:
\begin{equation}
\mathcal{L}_{\mathrm{LCM}}
= \mathbb{E}_{\tau,z_0,z_1}\Big[\big\|(z_1 - z_0) - v_\Theta(z_\tau,\tau,c_{\text{text}})\big\|_2^2\Big].
\end{equation}

\subsubsection{Subject Layout Guidance Condition}
\label{subsubsec:guidance}
To achieve precise control over multi-shot video generation, we introduce a 3D-aware subject layout control mechanism inspired by cinematic mise-en-scène, as shown in Fig.~\ref{fig:pipeline}. Formally, we define a storytelling video scene as a collection of shots $\mathcal{S} = \{S_1, S_2, \dots, S_K\}$. Within any given shot $S_k$, we represent $M$ active subjects as a set of time-varying, oriented 3D bounding boxes $\{B_{i,f}\}_{i=1}^M \in \mathbb{R}^{3 \times 8}$, where $f$ denotes the frame index. To capture fine-grained geometric properties beyond simple spatial occupancy, we apply a Normalized Object Coordinate Space (NOCS) encoding function $\Phi(\cdot)$ to these bounding boxes. This NOCS-based color encoding transforms abstract 3D rotations and local coordinate semantics into dense, orientation-aware visual cues, enabling accurate perception of intricate subject dynamics, such as a character turning or adjusting their posture in 3D space.

\paragraph{3D-to-2D Projection.} we bridge the modality gap between the 3D staging environment and the 2D video generation process. For a specific shot $S_k$, we project the NOCS-encoded 3D bounding boxes onto the corresponding 2D image plane $\Pi_k$, yielding a per-frame orientation map sequence:
\begin{equation}
    L_f = \bigcup_{i=1}^M \Pi_{k,f} \left( \Phi(B_{i,f}) \right).
\end{equation}
Our oriented 3D bounding boxes encode explicit geometry with NOCS, their 2D projections preserve subject orientation and spatial relationships, largely alleviating the ambiguity of conventional 2D boxes.
Because directly processing this layout sequence in pixel space introduces significant computational overhead and modality misalignment, we compress it into the latent space. We utilize the pre-trained 3D VAE encoder $\mathcal{E}_{\text{vae}}$ to extract the layout latent $z_{L_f} = \mathcal{E}_{\text{vae}}(L_f)$. This step effectively aligns the structured 3D layout constraints with the latent distribution natively processed by the backbone DiT.

\paragraph{Orientation Map Injection.} To enforce these spatial-temporal layout constraints during video synthesis, we design a Control DiT injection module following ControlNet~\cite{zhang2023adding}. Architecturally, we construct a control branch by cloning the first $N/2$ transformer blocks from the base DiT model, ensuring the control network inherits robust foundational feature representations. The layout latent $z_L$ is fed into this branch as the primary conditioning input. To integrate the control signals organically without disrupting the generative prior of the base model, we employ the additive injection strategy. Specifically, the hidden states $\Delta h_j$ produced by the $j$-th layer of the control network are modulated via zero-initialized convolutions and then added to the corresponding hidden states $h_j$ of the base model as:
\begin{equation}
    \hat{h}_j = h_j + \text{ZeroConv}(\Delta h_j).
\end{equation}

This hierarchical injection seamlessly embeds fine-grained 3D layout guidance into the denoising process, enabling controllable synthesis of subject positions, poses, and consistent motion trajectories across multiple shots.

\begin{figure}[t]
  \centering
  \includegraphics[width=0.95\linewidth]{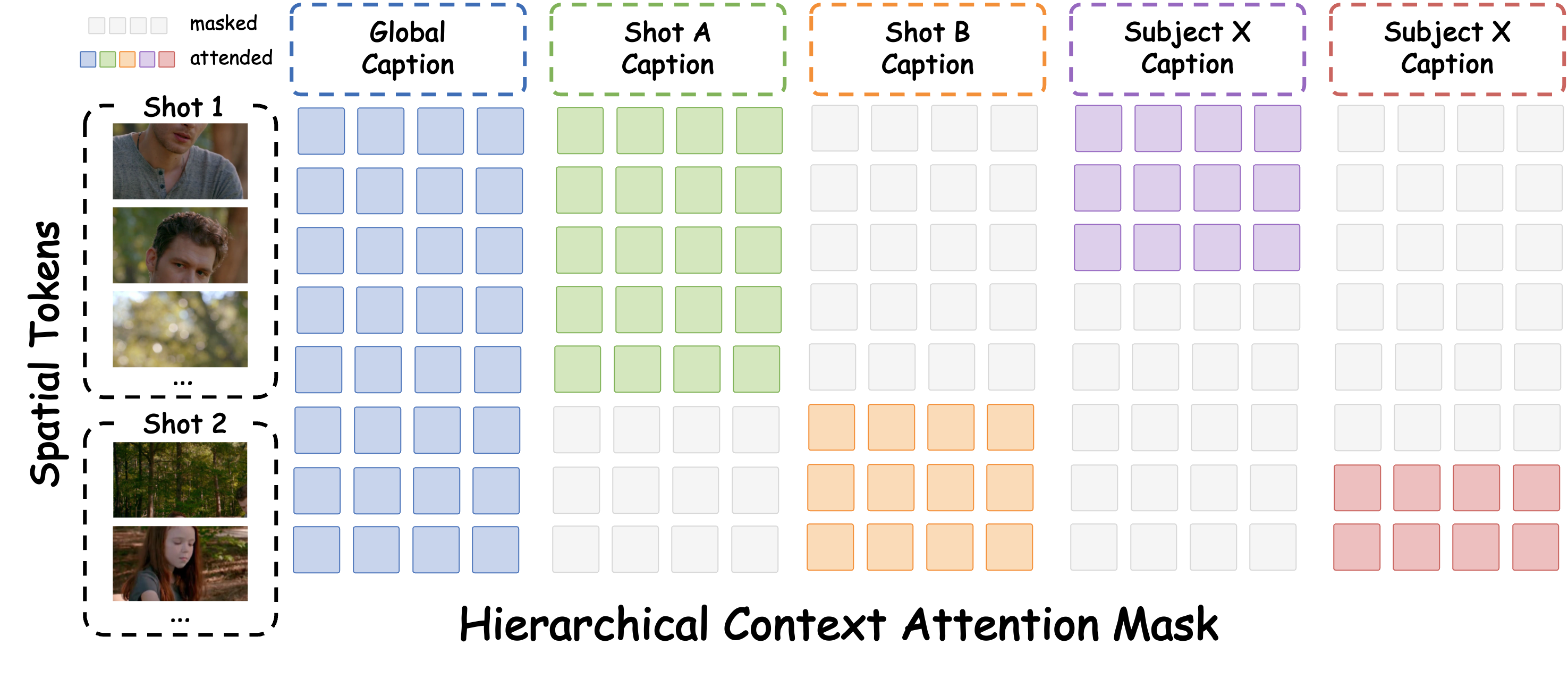}
  \caption{Visualization of the Hierarchical Context Attention Mask, where each visual token attends only to its corresponding text tokens via cross-attention.}
  \label{fig:hca_mask}
\end{figure}




\begin{figure*}[t]
  \centering
  \includegraphics[width=1.0\linewidth]{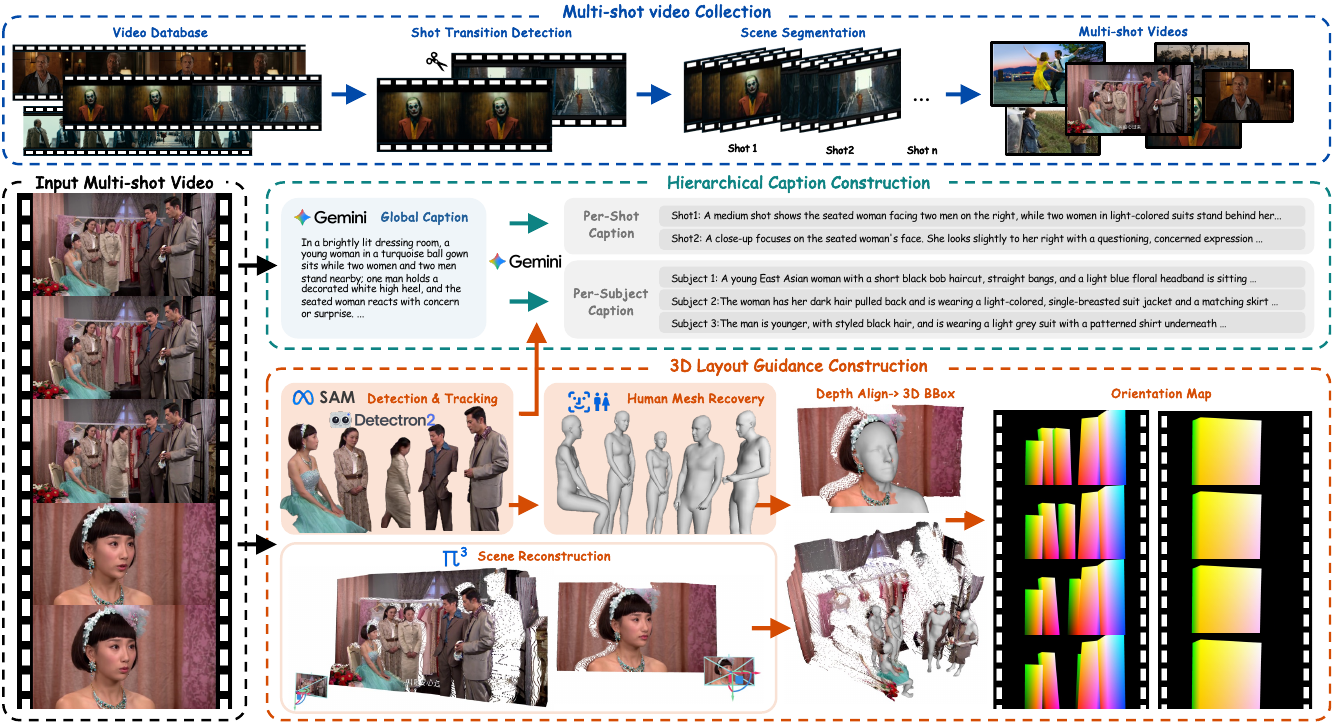}
  \caption{Multi-shot data annotation pipeline. 
  First, we collect videos from the Internet and segment them into shots using TransNetV2\cite{soucek2024transnet} to form multi-shot samples. Second, we construct 3D layouts by detecting and tracking subjects (Detectron2\cite{wu2019detectron2} + SAM2\cite{ravi2025sam}), recovering human meshes (CameraHMR\cite{patel2025camerahmr}), reconstructing scene depth and aligning meshes to a unified coordinate system (Pi3\cite{wang2025pi}), then encoding 3D boxes in NOCS and projecting them to 2D layout sequences. Third, we generate hierarchical captions (global/shot/subject) with Gemini, including consistent subject IDs across shots.}
  \label{fig:data_pipeline}
\end{figure*}

\subsubsection{Hierarchical Context Attention Mask}\label{subsubsec:hca}
In our framework, as shown in ``Hierarchical Caption'' of Fig.~\ref{fig:pipeline}, the complete textual prompt $T$ is decoupled into three semantic levels: a global caption $T_G$ dictating the overall style and narrative, a set of shot-level captions $\{T_{S_k}\}_{k=1}^K$ describing the specific background and action of each shot, and a set of subject-level captions $\{T_{O_{k,i}}\}_{i=1}^M$ detailing the fine-grained attributes of subject $O_i$ within shot $S_k$.

\paragraph{Hierarchical Context Attention.} 
While the 3D layout control explicitly defines the spatial-temporal dynamics of a subject, we must also ensure that the subject's appearance aligns strictly with its specific textual description, avoiding semantic bleed across different shots or subjects (e.g., a character's clothing style leaking into the background). 
To effectively associate the hierarchical text conditions with their corresponding visual regions, we propose a Hierarchical Context Attention (HCA) mechanism as shown in Fig.~\ref{fig:hca_mask},
where the subject regions are obtained by projecting the corresponding 3D bounding boxes onto the 2D image plane without requiring additional user specification.
Standard cross-attention globally broadcasts all text tokens to all visual tokens, which often leads to attribute confusion in complex multi-shot generation. 
To mitigate this, HCA modifies the attention weights by introducing a meticulously designed mask matrix that restricts the receptive field of visual tokens.

Let $X_k$ denote the visual query tokens belonging to shot $S_k$. Leveraging the 2D layout projections $\Pi_{k,f}(B_{i,f})$ defined in the previous section, we can dynamically identify the subset of visual tokens $X_{f,i} \subset X_f$ that fall within the spatial-temporal bounding box of subject $O_{f,i}$. For the cross-attention operation, let $Q$ denote the queries derived from the video tokens, and $K, V$ denote the keys and values derived from the hierarchical text $T$. We compute the masked cross-attention as:
\begin{equation}
    \text{Attention}(Q, K, V) = \text{Softmax}\left(\frac{QK^T}{\sqrt{d}} + \mathcal{M}\right)V,
\end{equation}
where $d$ is the dimension number, and $\mathcal{M}$ is the hierarchical attention mask matrix. For a specific visual query token $q \in X_k$ and a text key token $p \in T$, the mask value $\mathcal{M}_{q,p}$ is defined as follows:
\begin{equation}
    \mathcal{M}_{q,p} =
    \begin{cases}
    0, & \text{if } p \in T_G \\
    0, & \text{if } p \in T_{S_k} \\
    0, & \text{if } p \in T_{O_{k,i}} \text{ and } q \in X_{k,i} \\
    -\infty, & \text{otherwise}
    \end{cases}.
\end{equation}
By applying $0$ for valid semantic connections and $-\infty$ for masked regions, the exponentiation in the Softmax operation drives the attention weights of irrelevant text-visual pairs strictly to zero. Consequently, HCA ensures that global texts are visible to the entire video, shot-level texts only influence their respective shots, and subject-level texts are exclusively confined to the spatial bounds of the target subject. This explicit semantic binding efficiently captures multi-level information, significantly improving cross-shot coherence and fine-grained text-to-subject controllability.

\subsubsection{Training and Inference Paradigm}\label{subsubsec:training}
\methodname is trained in a two-stage manner. 
We first train the Hierarchical Context Attention Mask, which strengthens hierarchical context modeling and improves coherence across multi-shot sequences. 
We then introduce the Subject Layout injection module and continue training to endow the model with layout-conditioned, subject-level controllability. 
This second stage is conducted on our annotated dataset of 160k samples; throughout training, we randomly sample clips with different video lengths and varying numbers of shots to enhance robustness to diverse narrative structures.

During inference, the proposed framework enables controllable multi-shot video generation, maintaining text-driven inter-shot coherence as well as subject consistency. 
With explicit layout inputs, it further allows precise control over each subject’s attributes, including position and orientation. 
Moreover, both the number of shots and the duration of each shot are configurable, supporting highly customized multi-shot video storytelling.

\subsection{Multi-Shot Data Annotation}\label{subsec:annotation}
To support large-scale training, we build a scalable annotation pipeline tailored for multi-shot videos, consisting of three components: multi-shot video collection, 3D layout construction, and hierarchical caption generation (as shown in \figref{fig:data_pipeline}).

\subsubsection{Multi-shot video Collection.} At the data level, to obtain multi-shot videos at sufficient scale, we collect video clips from the Internet, covering sources such as TV series, movies, and documentaries. 
Since our method requires precise control at the shot level, we apply TransNet V2~\cite{soucek2024transnet} to the raw videos for shot boundary detection, producing scene/shot-level segments, based on which we construct multi-shot video samples.

\subsubsection{3D Layout Construction.}
To obtain a 3D layout representation analogous to a “shot blocking sheet,” we perform subject-level geometric construction. 
Concretely, we first use Detectron2~\cite{wu2019detectron2} to detect subjects in each shot, and then use SAM2~\cite{ravi2025sam} to track them across frames, obtaining a per-frame mask for each subject.
Next, we apply CameraHMR~\cite{patel2025camerahmr} for human mesh recovery to obtain SMPL-based 3D human representations~\cite{loper2023smpl}. 
To unify 3D geometric relationships across different subjects and camera viewpoints, we further use Pi3~\cite{wang2025pi} to reconstruct the entire scene and recover per-frame depth, and align the SMPL meshes with the depth maps, resulting in subject meshes in a unified scale and coordinate system. 
Finally, we compute a 3D bounding box that tightly encloses each mesh. Inspired by the NOCS~\cite{wang2019normalized, zhang2026unipr}, we encode the box in the Normalized Object Coordinate Space (NOCS), which preserves fine-grained geometric information such as orientation. We then project the NOCS-encoded 3D boxes onto the image plane, yielding a 2D layout sequence as the conditioning signal.

\subsubsection{Hierarchical Caption Construction.}
At the text annotation level, to improve controllability of text conditions and reduce semantic drift during generation, we construct hierarchical captions with three levels: global, shot, and subject. 
For each multi-shot video, we first use Gemini 3 Flash to generate a global caption for the entire video, focusing on the overall scene and major events; based on this, we further infer shot-level captions conditioned on the content of each shot.
For subject-level captions, we first provide Gemini with subject crops from each shot obtained via SAM2\cite{ravi2025sam}, and ask it to merge the same subject across the whole video and assign a globally unique subject ID. 
We then generate a description for each subject conditioned on the global caption, resulting in hierarchical text annotations aligned with the multi-shot structure and suitable for enforcing subject consistency.

\section{Experiments}


\begin{table*}[!t]
\centering

\caption{Quantitative results for multi-shot video generation. 
Among the 1B models, the best results are highlighted in \textbf{boldface}, while the second best are \underline{underlined}. 
Here, Sub., Bg., Cons., and Align. denote Subject, Background, Consistency, and Alignment, respectively.}


\resizebox{\linewidth}{!}{
\begin{tabular}{l|ccc|cc|cccc}
\toprule

\multirow{2}{*}{Method}
& \multicolumn{3}{c|}{Inter-shot Cons.}
& \multicolumn{2}{c|}{Intra-shot Cons.}
& \multirow{2}{*}[-2pt]{\makecell[c]{Text \\ Align. $\uparrow$}}
& \multirow{2}{*}[-2pt]{\makecell[c]{Trans. \\ Control $\uparrow$}}
& \multirow{2}{*}[-2pt]{\makecell[c]{Aesthetic \\ Quality $\uparrow$}}
& \multirow{2}{*}[-2pt]{\makecell[c]{Narrative \\ Coherence $\uparrow$}} \\

\cmidrule(lr){2-6}

& Semantic $\uparrow$
& Subject $\uparrow$
& Background $\uparrow$
& Subject $\uparrow$
& Background $\uparrow$
& & & & \\

\midrule

{HoloCine (14B)}
&{0.6423}
&{0.5955}
&{0.5385}
&{0.8101}
&{0.7099}
&{0.1986}
&{0.9826}
&{0.5598}
&{0.8839} \\

\midrule

EchoShot
& 0.5829
& 0.4492
& 0.4070
& 0.6717
& \underline{0.6251}
& \underline{0.1987}
& 0.9321
& 0.5414
& 0.8393 \\

MultiShotMaster
& \underline{0.5973}
& \underline{0.4751}
& \underline{0.5220}
& \underline{0.7150}
& 0.5962
& 0.1914
& \underline{0.9563}
& \underline{0.5452}
& \underline{0.8810} \\

\textbf{Ours}
& \textbf{0.6175}
& \textbf{0.5499}
& \textbf{0.5762}
& \textbf{0.7630}
& \textbf{0.6844}
& \textbf{0.1991}
& \textbf{0.9812}
& \textbf{0.5459}
& \textbf{0.8869} \\

\bottomrule
\end{tabular}
}

\label{tab:visual_results}
\end{table*}

\begin{table}[htbp]
  \centering
  \caption{Quantitative Comparison of Orientation Estimation Results}
  \label{tab:orientation_results}
  \resizebox{\linewidth}{!}{ 
  \begin{tabular}{l|ccc|cc}
    \toprule
    \multirow{2}{*}{Method} & 
    \multicolumn{3}{c|}{Mean Diff (rad)} & 
    \multirow{2}{*}{OS Yaw $\uparrow$} &
    \multirow{2}{*}{mAOE (rad) $\downarrow$}\\
    \cmidrule(lr){2-4} 
    & Yaw $\downarrow$ & Pitch $\downarrow$ & Roll $\downarrow$ & & \\
    \midrule
    MultiShotMaster & \underline{0.3188} & 0.5933 & \underline{0.3160} & \underline{0.9222} & 0.4094 \\
    EchoShot        & 0.3690 & \underline{0.4304} & 0.4241 & 0.8963 & \underline{0.4079} \\
    Ours            & \textbf{0.1497} & \textbf{0.2641} & \textbf{0.1909} & \textbf{0.9641} & 
    \textbf{0.2015} \\
    \bottomrule
  \end{tabular}
  }
\end{table}

\begin{table}[htbp]
  \centering
  \caption{Ablation Study on 3D Layout Guidance}
  \label{tab:ablation_results_orientation}
  \resizebox{\linewidth}{!}{ 
  \begin{tabular}{l|ccc|cc}
    \toprule
    \multirow{2}{*}{Method} & 
    \multicolumn{3}{c|}{Mean Diff (rad)} & 
    \multirow{2}{*}{OS Yaw $\uparrow$} &
    \multirow{2}{*}{mAOE (rad) $\downarrow$}\\
    \cmidrule(lr){2-4} 
    & Yaw $\downarrow$ & Pitch $\downarrow$ & Roll $\downarrow$ & & \\
    \midrule
    w.o. 3D Layout   & 0.4983 & \textbf{0.2470} & 0.5130 & 0.8575 & 0.4194 \\
    Base             & \textbf{0.1497} & 0.2641 & \textbf{0.1909} & \textbf{0.9641} & \textbf{0.2015} \\
    \bottomrule
  \end{tabular}
  }
\end{table}

\begin{table}[htbp]
  \centering
  \caption{Ablation Study on Hierarchical Context Attention}
  \label{tab:ablation_results_hca}
  \resizebox{\linewidth}{!}{ 
  \begin{tabular}{l|ccc|ccc}
    \toprule
    \multirow{2}{*}{Method} 
    & \multicolumn{3}{c|}{Inter-shot Cons.} 
    & \multirow{2}{*}[-2pt]{\makecell[c]{Text \\ Align. }$\uparrow$}
    & \multirow{2}{*}[-2pt]{\makecell[c]{Trans. \\ Control }$\uparrow$}
    & \multirow{2}{*}[-2pt]{\makecell[c]{Narrative \\ Coherence }$\uparrow$} \\
    \cmidrule(lr){2-4} 
    & Semantic $\uparrow$ & Sub. $\uparrow$ & Bg. $\uparrow$ & & & \\
    \midrule
    w.o. Sub       & \underline{0.6098} & 0.5468  & 0.5239  & \textbf{0.2001}  & \textbf{0.9858}  & \underline{0.8452} \\
    w.o. Mask  & 0.3560   & \textbf{0.7781}  & \underline{0.6014}  & 0.1530  & 0.9461  & 0.8304 \\
    Base      & \textbf{0.6180}   & \underline{0.5905}  & \textbf{0.6166}  & \underline{0.1969}  & \underline{0.9771}  & \textbf{0.8631} \\
    \bottomrule
  \end{tabular}
  }
\end{table}

\subsection{Implementation Details.}
Our model is built upon a 1B-parameter single-shot text-to-video (T2V) backbone and is trained in two stages. 
In the first stage, we train the single-shot model with our Hierarchical Captions to enable precise shot-level control and fine-grained subject description. 
In the second stage, we extend controllability with 3D layout guidance by training only the ControlDiT branch while keeping the rest of the model frozen. 
We annotate 160K multi-shot videos, each comprising 2–6 shots with up to 250 frames. 
All experiments are conducted on 32 GPUs with a batch size of 1.

\subsection{Evaluation}

\paragraph{Dataset.} 
Following prior work\cite{wang2025multishotmaster, luo2026shotstream, meng2025holocine}, we use Gemini 2.5 to design 100 samples of multi-shot prompts, and adapt each prompt to match the required input style of each baseline method. 
These prompts cover a wide range of topics, enabling a comprehensive assessment of a model’s ability to maintain consistency across diverse scenarios in multi-shot narrative video generation. 

\paragraph{Metrics.} 
We evaluate performance from multiple aspects: 
(1) Text Alignment: following prior work~\cite{wang2025multishotmaster, luo2026shotstream}, we compute the similarity between text features and per-shot video features extracted by ViCLIP\cite{wang2024internvid}. 
(2) Inter-shot Consistency: we first measure the holistic semantic similarity between ViCLIP shot features; then, on keyframes (first, middle, and last frames), we apply YOLOv11\cite{khanam2024yolov11} and ByteTrack\cite{zhang2022bytetrack} to detect and crop subjects and backgrounds, and use DINOv2\cite{oquab2023dinov2} feature to quantify subject consistency and scene consistency across shots. 
(3) Intra-shot Consistency: within each shot, we similarly use YOLOv11 and ByteTrack to crop subjects and backgrounds from keyframes, and employ DINOv2 feature to evaluate subject consistency and scene consistency within the shot. 
(4) Transition Control: following prior work\cite{meng2025holocine}, we report the shot cut accuracy score to measure shot-cut accuracy. 
(5) Overall Quality: following prior work\cite{luo2026shotstream, wang2025multishotmaster}, we report Aesthetic Quality and the Narrative Coherence score to assess overall narrative quality.
(6) Orientation Quality: We assess controllability by comparing subject orientations in the generated videos and GT, reporting yaw/pitch/roll errors as well as mAOE and OSYaw.

\subsubsection{{Quantitative Results}}
\tabref{tab:visual_results} reports quantitative comparisons with EchoShot~\cite{wang2025echoshot} and MultiShotMaster~\cite{wang2025multishotmaster} on multi-shot video generation.
Overall, our approach achieves the best performance across all metrics, demonstrating improved visual consistency, text alignment, and narrative quality for multi-shot video generation.
For consistency, we outperform prior methods on both intra-shot and inter-shot Subject/Background scores, indicating more stable subjects and backgrounds within each shot and more reliable identity/scene continuity across shot transitions. 
This gain mainly comes from the Hierarchical Context Attention Mask Mechanism, which enforces constraints over shots and subjects, and the injected 3D layout, which provides an explicit spatial-structure prior.
Moreover, our method ranks first in Text Alignment and Narrative Coherence, suggesting better adherence to the text prompt and stronger multi-shot storytelling coherence.

\tabref{tab:orientation_results} compares the subject-orientation errors of videos generated by different methods. 
Our method achieves substantially lower Mean Diff on yaw/pitch/roll and lower mAOE than MultiShotMaster and EchoShot, while obtaining the highest OS Yaw. 
These results indicate that the subject orientation in our generated videos is closer to the target pose specified by the storyboard. 
This validates our shot-level controllability under layout conditions, enabling users to define the storyboard and layout and then generate fine-grained multi-shot videos matching the intended shot plan.

\subsubsection{{Qualitative Results}}
\begin{figure*}[t]
    \centering
    \includegraphics[width=0.9\linewidth]{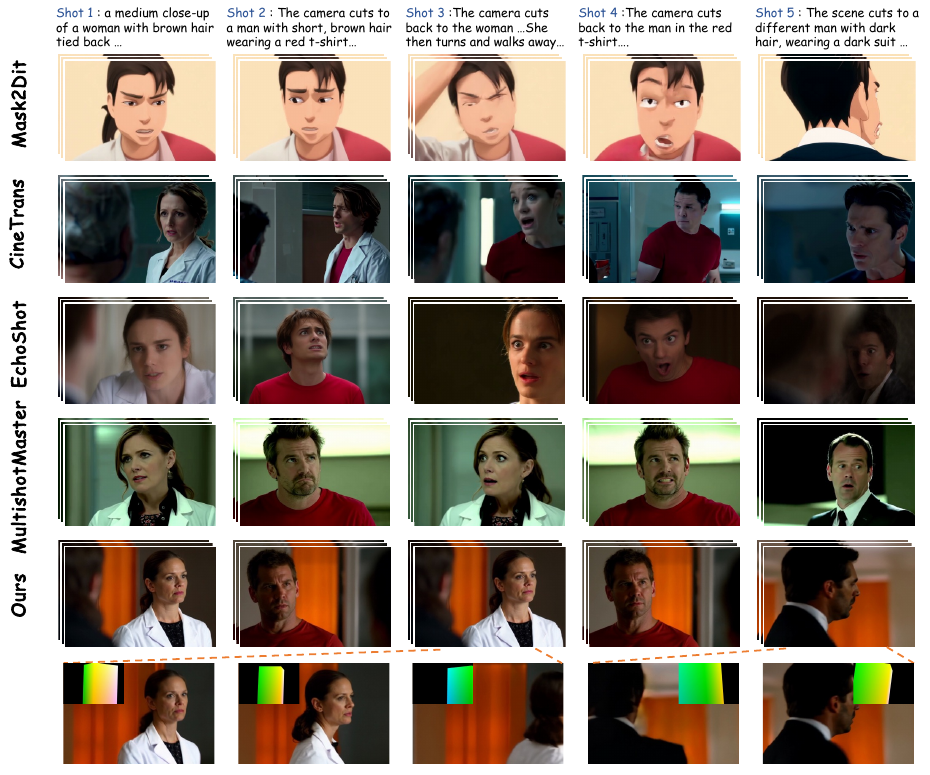}
    \caption{Qualitative comparison. We show frames from each shot generated by all compared methods. Our method not only strictly follows the prompt and maintains strong visual coherence, but also produces natural transitions between shots. In addition, our method closely follows the guidance provided by the 3D layout, especially in Shots 3 and 5, where it better realizes the subject's layout and motion.}
    \label{fig:compare}
\end{figure*}

\begin{figure*}[t]
    \centering
    \includegraphics[width=1\linewidth]{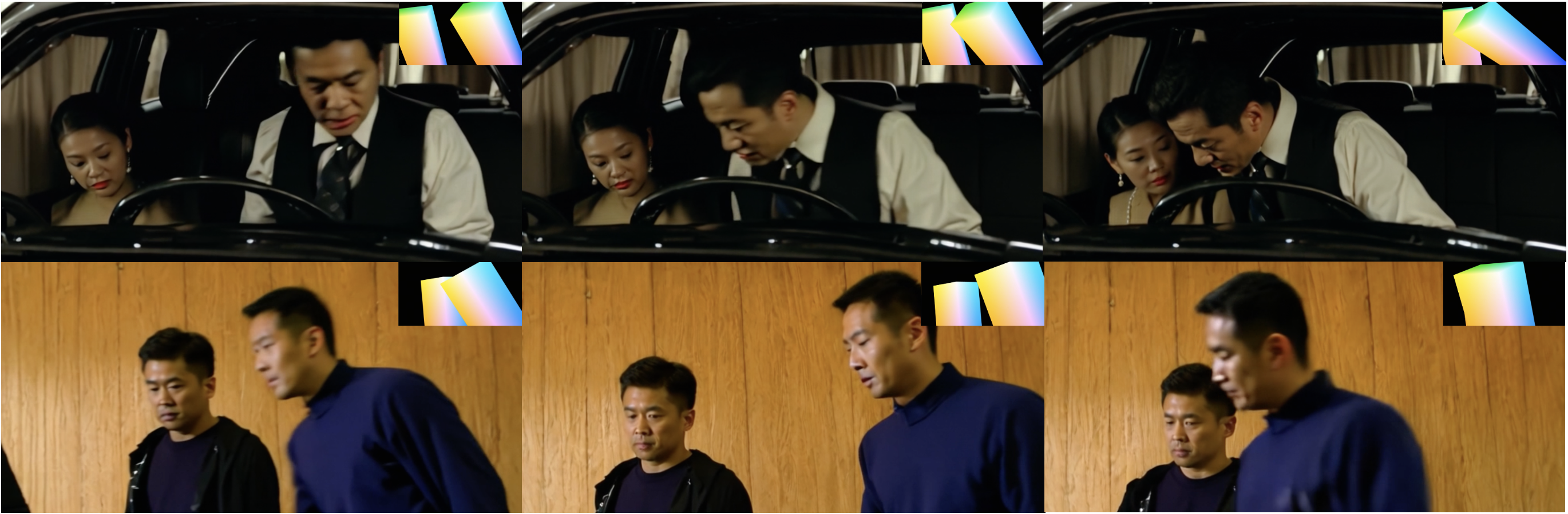}
    \caption{Qualitative demonstration of controlling the subject dynamics using our orientation maps.}
    \label{fig:control_orientation_map}
\end{figure*}

\begin{figure*}[t]
    \centering
    \includegraphics[width=\linewidth]{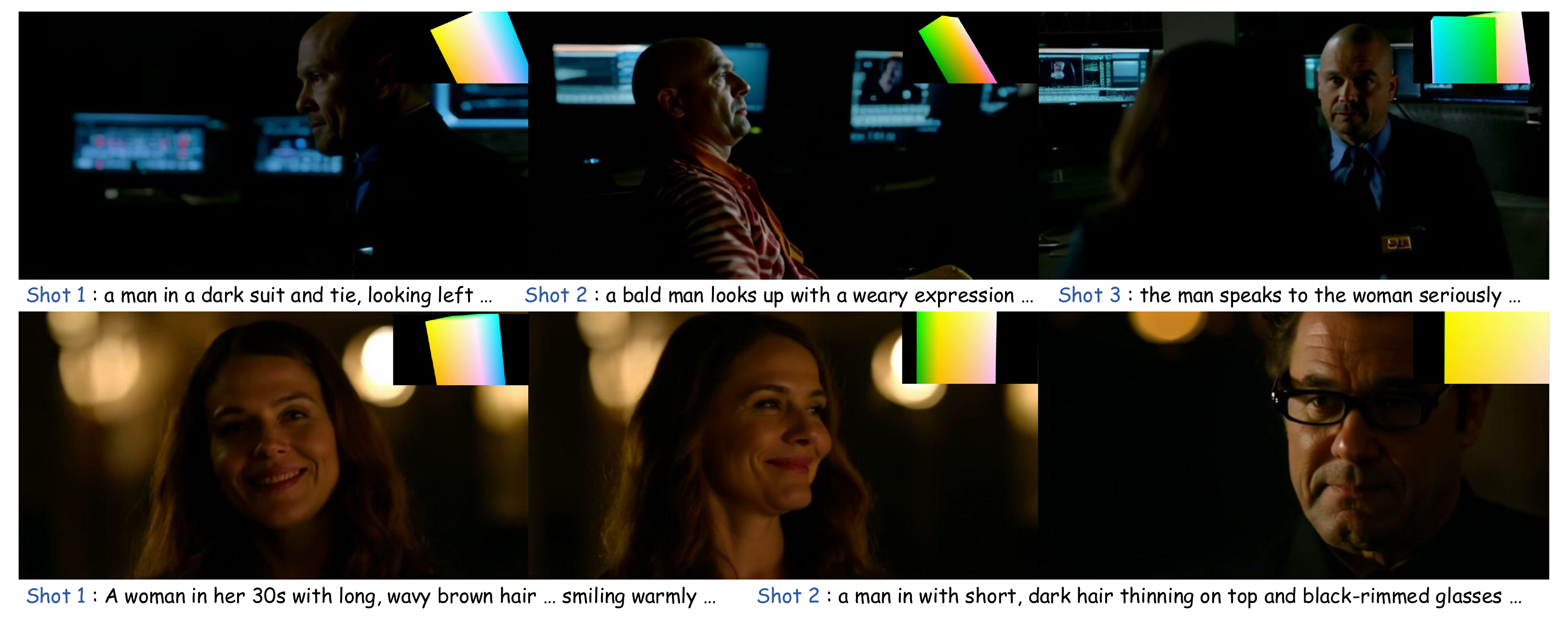}
    \caption{More qualitative results of \methodname on multi-shot video generation.}
    \label{fig:more_results}
\end{figure*}

As shown in \figref{fig:compare}, we compare our method with EchoShot and MultiShotMaster. 
EchoShot struggles to maintain consistent scenes and subjects under cross-view generation, and its outputs do not align well with the corresponding prompts. 
MultiShotMaster can preserve subject consistency reasonably well across viewpoints, but it offers limited control over the subject’s motion, making it difficult to precisely satisfy the actions and trajectories required by the storyboard. 
In contrast, our method maintains high visual quality while preserving both scene/background and subject consistency across shots, and it enables stronger, fine-grained control over the subject’s motion, making it well-suited for storyboard-driven multi-shot video generation. \figref{fig:more_results} shows more multi-shot results while \figref{fig:control_orientation_map} shows results on how \methodname control dynamics with 3D layout.

\subsection{Ablation Study}
\subsubsection{Ablation Study on Hierarchical Context Attention}
\tabref{tab:ablation_results_hca} reports an ablation study on Hierarchical Context Attention to assess its contribution to multi-shot generation. The full model (Base) delivers the best overall trade-off across key metrics.
(1) Notably, removing the mask (w.o. Mask) increases subject and background consistency, but substantially degrades semantic consistency and text alignment, and also weakens transition control and narrative coherence. 
This indicates that without the mask constraint, the model tends to collapse multi-shot generation into a near single continuous video, yielding superficially consistent subjects/backgrounds but failing to follow shot-wise semantic changes and textual conditions. 
(2) Removing the subject branch (w.o. Sub) makes the model more compliant in text alignment and transition control, yet it significantly harms cross-shot subject consistency and consequently reduces narrative coherence, highlighting the importance of subject-level hierarchical context for preserving identity across shots. 
Overall, these results validate that Hierarchical Context Attention improves cross-shot coherence, text adherence, and controllable transitions, leading to more coherent multi-shot narratives.

\subsubsection{Ablation Study on 3D Layout Guidance}
Table \ref{tab:ablation_results_orientation} reports the ablation study to assess the impact of 3D layout guidance. 
The results show that incorporating 3D layout constraints substantially reduces the overall orientation error (Mean Diff and mAOE) and notably improves yaw alignment (OS Yaw), indicating that the generated camera/subject orientations reliably follow the intended 3D shot plan. 
This improvement better matches a practical filmmaking workflow, where a director specifies storyboarded shots and camera blocking, and the system “executes” the plan like a cinematographer. 
Importantly, the gain on yaw is particularly significant: in typical cinematography, yaw (horizontal heading) largely determines key narrative cues such as facing direction, eyeline relationships, and screen-space spatial continuity, whereas pitch often varies less and is generally less decisive for shot-to-shot staging. 
Overall, this validates that 3D layout guidance enhances controllability and predictability of shot orientation, enabling high-fidelity storyboard-driven generation.

\subsection{Application: Versatile Creative Control}

\begin{figure*}[t]
    \centering
    \includegraphics[width=1.0\linewidth]{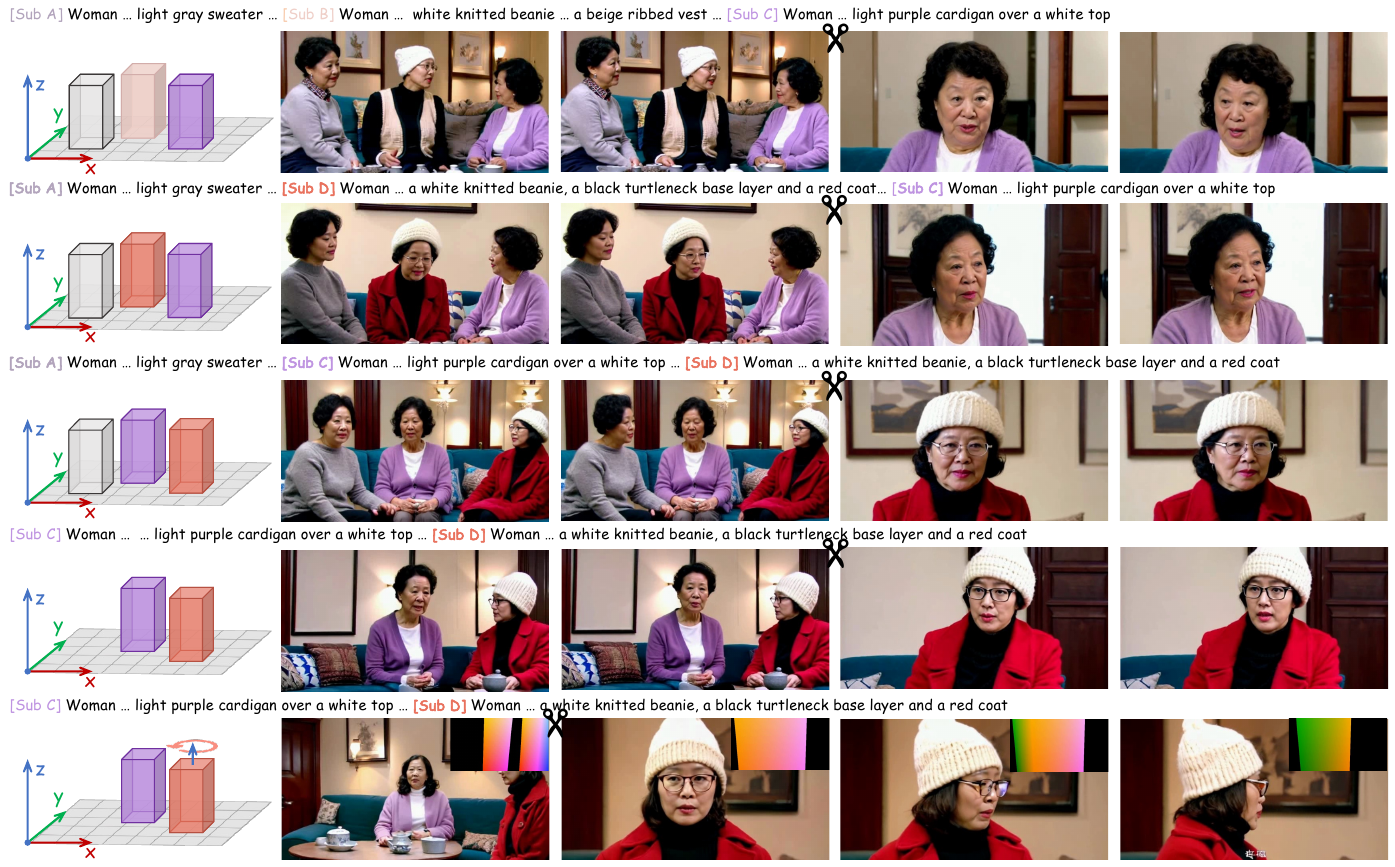}
    \caption{\methodname enables versatile control over the generated multi-shot videos. (Row 1$\to$2) We change the textual descriptions while keeping the layout unchanged. Note we change from Subject B with ``white knitted beanie'' to Subject D ``red coat''. (Row 2$\to$3) We keep the text unchanged and swap the positions of Subject C and Subject D. (Row 3$\to$4) We delete Subject A. (Row 4$\to$5) We let Subject A turn to the right in another shot. (Note that different rows are generated separately. Even though different rows use the same identity text description, the identity consistency is not ensured.)}
    \label{fig:edit}
\end{figure*}

To demonstrate the versatile controllability, we further show the generation results in Fig.~\ref{fig:edit}, where \methodname enables creators to generate diverse visual outcomes from a unified 3D layout by varying the textual descriptions. At the same time, our method can also vary spatial configurations with the same identity descriptions. This decoupling of 3D spatial layout and semantic descriptions empowers creators with steerability in multi-shot video production.



\section{Conclusions}
In this work, we propose \methodname, a controllable multi-shot video generation framework with both cross-shot consistency and fine-grained controllability. 
The key idea is to adopt a 3D-aware layout guidance and a hierarchical context attention mechanism to enable the 3D staging in the multi-shot video generation.
We further develop an automatic data curation and annotation pipeline and construct a 160K multi-shot dataset for training.
We hope this work will inspire future research on production-oriented, 3D-aware controllable multi-shot video generation.

\paragraph{Limitation and Future} 
While \methodname demonstrates strong controllability and consistency, its overall generation quality is still limited by the capacity of the pretrained backbone. 
With a relatively constrained parameter scale, the model may struggle to generate high-quality dynamic videos.
In future work, we plan to scale up the model capacity by adopting larger backbones and/or increasing the parameter budget of the controllable modules.


\vspace{2.0cm}


{
    \small
    \bibliographystyle{ieeenat_fullname}
    \bibliography{main}
}


\end{document}